\documentclass[lettersize,journal]{IEEEtran}

\usepackage{float}
\usepackage{stfloats}
\usepackage{cite}

\usepackage{multirow}
\usepackage{booktabs} % for better table rules
\usepackage{amsmath,amsfonts}
\usepackage{algorithmic}
\usepackage{array}
\usepackage[caption=false,font=normalsize,labelfont=sf,textfont=sf]{subfig}
\usepackage{textcomp}
\usepackage{stfloats}
\usepackage{url}
\usepackage{verbatim}
\usepackage{graphicx}
\def\BibTeX{{\rm B\kern-.05em{\sc i\kern-.025em b}\kern-.08em
    T\kern-.1667em\lower.7ex\hbox{E}\kern-.125emX}}
\usepackage{balance}

\begin{document}
\title{A novel seamless magnetic-based actuating mechanism for end-effector-based robotic rehabilitation platforms}
\author{Sima~Ghafoori\textbf{\large{*}},
        Ali~Rabiee\textbf{\large{*}},
        Maryam Norouzi,
        % Yalda Shahriari,
        Musa~Jouaneh
        and~Reza~Abiri,~\IEEEmembership{~IEEE}% <-this % stops a space
        
\thanks{S. Ghafoori, A. Rabiee, and R. Abiri were with the Department
of Electrical, Computer, and Biomedical Engineering, University of Rhode Island, Kingston,
RI, 02881 USA e-mail: (reza\_abiri@uri.edu).}% <-this % stops a space
\thanks{M. Norouzi and M. Jouaneh were with the Department of Mechanical Engineering, University of Rhode Island, Kingston, RI, 02881 USA}% <-this % stops a space
\thanks{\textbf{\large{*}}These authors contributed equally to this work}
}

\maketitle

\begin{abstract}
Rehabilitation robotics continues to confront substantial challenges, particularly in achieving smooth, safe, and intuitive human-robot interactions for upper limb motor training. Many current systems depend on complex mechanical designs, direct physical contact, and multiple sensors, which not only elevate costs but also reduce accessibility. Additionally, delivering seamless weight compensation and precise motion tracking remains a highly complex undertaking. To overcome these obstacles, we have developed a novel magnetic-based actuation mechanism for end-effector robotic rehabilitation. This innovative approach enables smooth, non-contact force transmission, significantly enhancing patient safety and comfort during upper limb training. To ensure consistent performance, we integrated an Extended Kalman Filter (EKF) alongside a controller for real-time position tracking, allowing the system to maintain high accuracy or recover even in the event of sensor malfunction or failure. In a user study with 12 participants, 75\% rated the system highly for its smoothness, while 66.7\% commended its safety and effective weight compensation. The EKF demonstrated precise tracking performance, with root mean square error (RMSE) values remaining within acceptable limits (under 2 cm). By combining magnetic actuation with advanced closed-loop control algorithms, this system marks a significant advancement in the field of upper limb rehabilitation robotics.

\end{abstract}


\section{Introduction}

\IEEEPARstart{R}{ehabilitation} robots are becoming essential tools for helping individuals recover from sensorimotor impairments, particularly those caused by strokes \cite{anwer2022rehabilitation}. Traditional physical therapy often struggles to meet the intensive demands of the expanding stroke patient population \cite{khalid2023robotic}. To address this, robotic systems for upper limb rehabilitation have been developed to assist and accelerate recovery by enabling patients to practice and strengthen their motor abilities independently \cite{maciejasz2014survey, demofonti2021affordable, narayan2021development, johnson2017affordable}. These systems have demonstrated promising results, performing comparably to conventional therapies in several studies \cite{khalid2023robotic, mansour2022efficacy}. However, challenges remain, particularly in developing actuation mechanisms that provide smooth and controlled movement \cite{maciejasz2014survey}. Magnetic technology offers a promising solution by enabling motion transmission without physical contact, allowing for precise position tracking even when objects are out of direct view \cite{abbott2020magnetic, LI2018278, 6631346}. While magnetic technology has been widely applied in medical fields such as localization, actuation, screening, and drug delivery \cite{abbott2020magnetic, LI2018278, 6631346}, its use in rehabilitation robotics remains relatively limited \cite{Chaloupka2002, brainsci12010113, abbott2020magnetic}. This technology holds the potential to simplify robot design while enhancing the efficiency and safety of upper limb rehabilitation platforms.

Upper limb rehabilitation robots are typically categorized as end-effector- or exoskeleton-based \cite{narayan2021development}. While exoskeleton robots offer a broader range of motion due to multiple connection points \cite{narayan2021development}, end-effector robots are more popular because of their simplicity, ease of production, control, and adaptability to different arm sizes \cite{maciejasz2014survey}. These robots are also more cost-effective and have greater commercialization potential \cite{maciejasz2014survey}, with studies showing their effectiveness over exoskeletons in certain applications \cite{lee2020comparisons}. End-effector robots can be further classified into systems for manipulation, reaching, or both \cite{maciejasz2014survey, campolo2014h, ceccarelli2021operation, wu2019adaptive, guillen2021usability, qian2021quantitative, jandaghi2024composite, jandaghi2024adaptive}. Reaching robots, which are often prioritized for training upper extremity movements, can be adapted to provide between one and three degrees of freedom, enabling targeted movement training across multiple axes \cite{maciejasz2014survey, basteris2014training}. In improving control accuracy, the Kalman Filter plays a critical role, particularly in robotic rehabilitation systems. Widely used for disturbance estimation, such as in exoskeletons like RehabRoby \cite{ozkul2011design}, and for sensor fusion to minimize the number of sensors without sacrificing control quality \cite{sun2021sensor}, it also enhances joint position estimation in systems like Kinect \cite{das2018improving} and improves motion tracking in virtual reality rehabilitation setups \cite{kim2013development}.

Although planar upper limb rehabilitation platforms showed promising results in assisting occupational therapies and associated patients in gaining higher clinical outcomes \cite{maciejasz2014survey, basteris2014training}, there are still several challenges that should be addressed by innovative approaches to offer portable, safer, and low-cost platforms. Such a platform should ensure user safety during direct interaction \cite{wu2019adaptive, ellis2007act}, compensate for the weight of the human hand while maintaining smooth motion \cite{wu2019adaptive, zhang2024research}, and improve accessibility for home-based rehabilitation \cite{maciejasz2014survey}. In response to these challenges, we developed a novel planar robotic rehabilitation platform that leverages magnetic technology and nonlinear dynamical modeling. To the best of our knowledge, this is the first study to apply magnetic actuation in end-effector-based upper limb rehabilitation robots. Our system’s end-effector compensates for the user’s hand weight, ensuring smoother, more natural motion while providing seamless force transmission from the motors to the hand and wrist. For improved safety, the platform is enclosed in a plexiglass box, maintaining a secure distance between the user and the system’s mechanical components. Additionally, to enhance control accuracy and motion estimation, we integrated an Extended Kalman Filter into the system, allowing for precise disturbance estimation and improved control performance. This combination of magnetic actuation and advanced control strategies results in a simple, cost-effective solution for safer and more efficient upper limb rehabilitation. By focusing on a one-dimensional design, we maintained affordability and simplicity, making the system more accessible for widespread adoption in home-based settings. Our specific contributions are as follows: (1) Developed an innovative magnetic-based actuation mechanism for non-contact, smooth, and safe haptic force transmission. (2) Derived nonlinear dynamic models that integrate both frictional and magnetostatic forces, enabling precise control. (3) Enhanced the user interaction and interface, as validated by a user study showing high satisfaction ratings for smoothness, safety, and effective weight compensation.

\setlength{\textfloatsep}{0.2em}  % Adjust space between figure and text

\begin{figure}[t]
    \centering
    % Row 1
    \begin{minipage}[b]{0.15\textwidth}
        \centering
        \includegraphics[width=\linewidth]{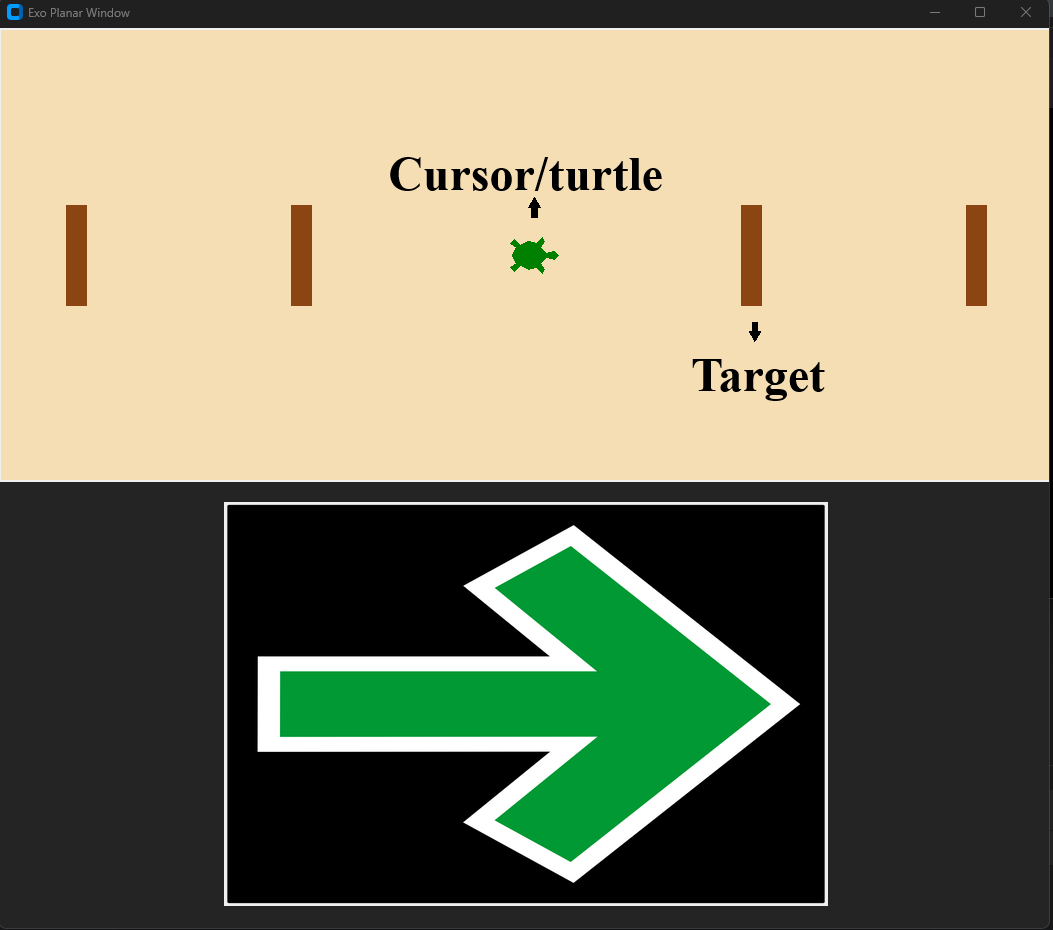}
        \captionsetup{justification=raggedright,singlelinecheck=false}
        \centerline{(a)}
    \end{minipage}
    \hfill
    \begin{minipage}[b]{0.33\textwidth}
        \centering
        \includegraphics[width=\linewidth]{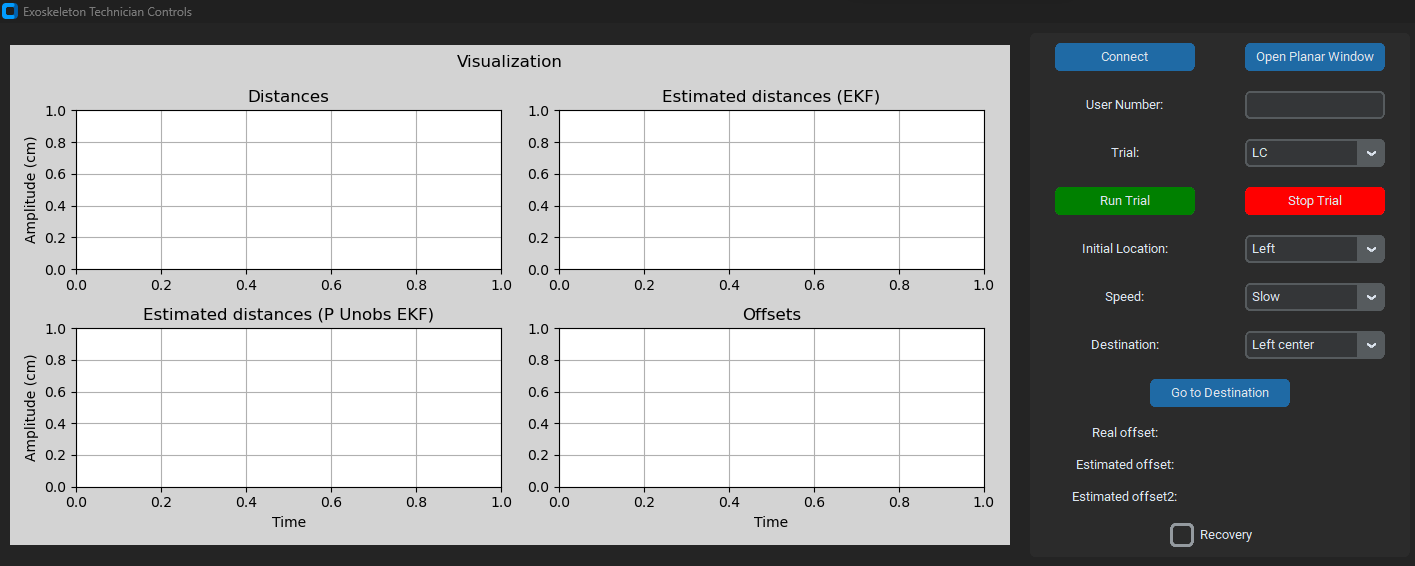}
        \captionsetup{justification=raggedright,singlelinecheck=false}
        \centerline{(b)}
    \end{minipage}
    \vspace{0.2em} % Adds space between the rows
    % Row 2
    \begin{minipage}[b]{0.48\textwidth}
        \centering
        \includegraphics[width=\linewidth]{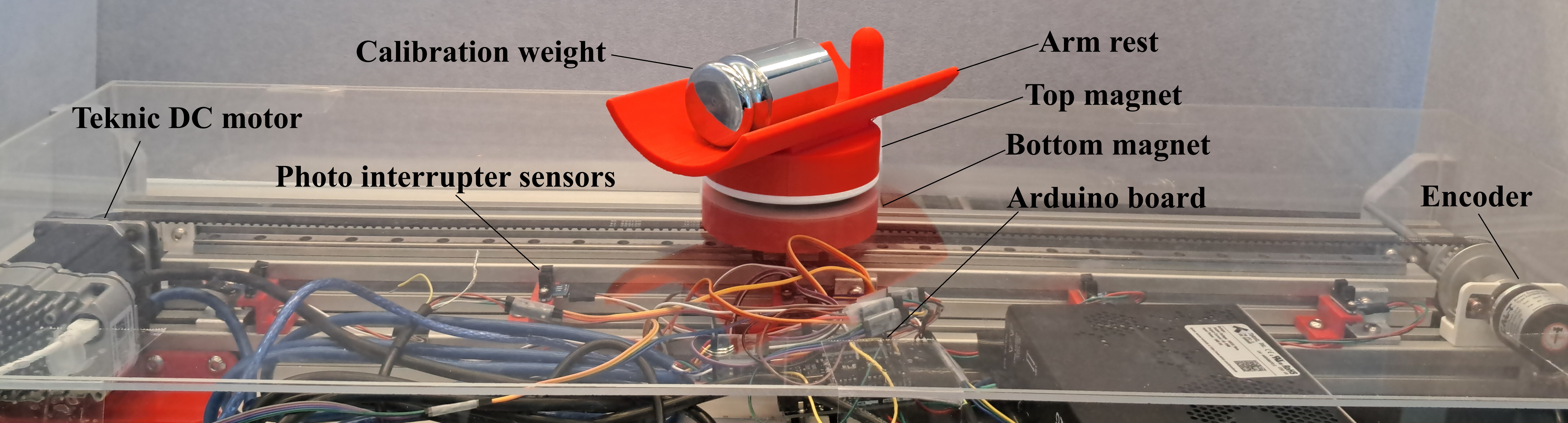}
        \captionsetup{justification=raggedright,singlelinecheck=false}
        \centerline{(c)}
    \end{minipage}
    \begin{minipage}[b]{0.48\textwidth}
        \centering
        \includegraphics[width=\linewidth]{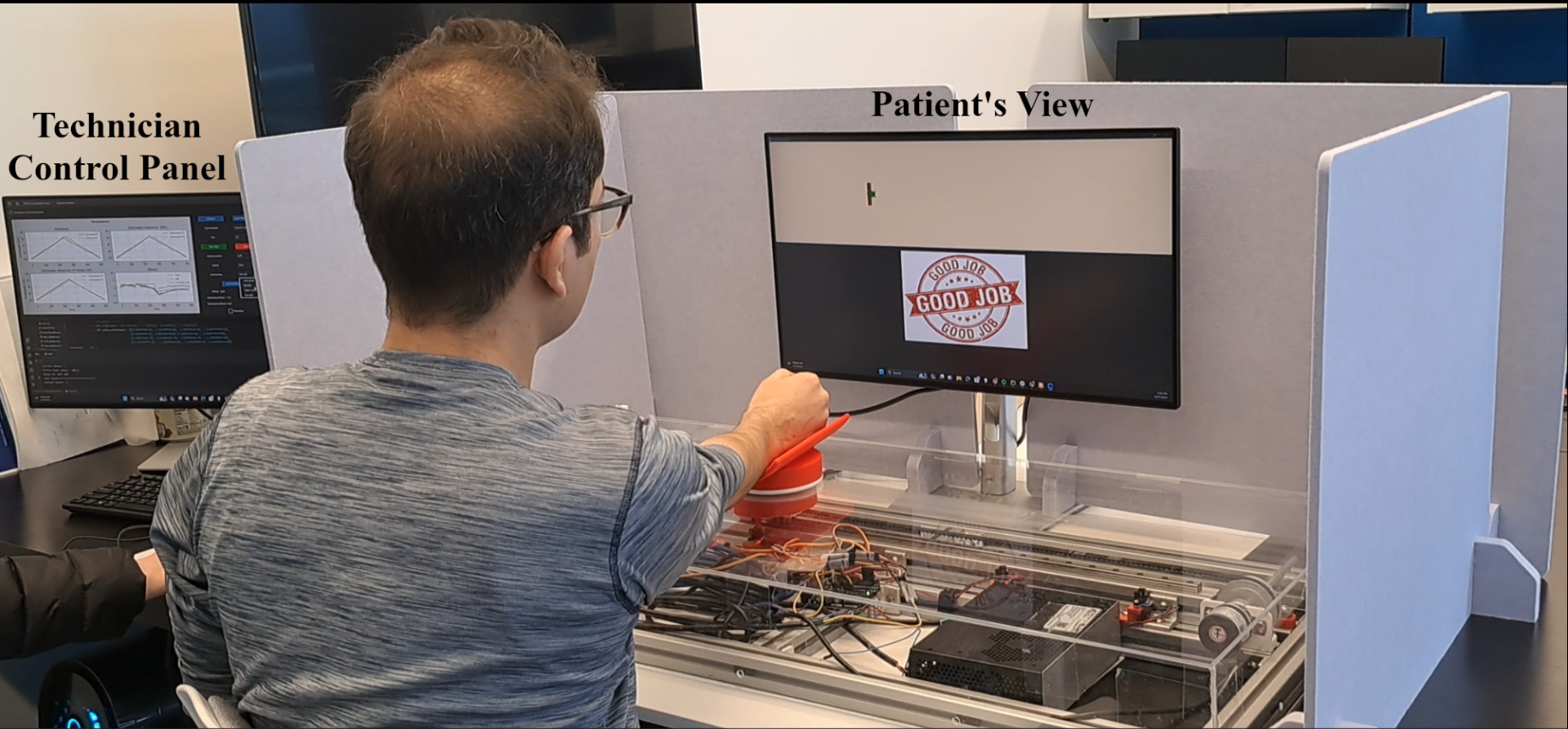}
        \captionsetup{justification=raggedright,singlelinecheck=false}
        \centerline{(d)}
    \end{minipage}
    \caption{The designed Graphical User Interface (GUI): a) patient’s view, b) The technician control panel. c) The designed planar robotic platform. d) a preview of running a trial with a participant}
    \label{hardware}
\end{figure}

\begin{figure}[t]
\centering

\begin{minipage}{.48\textwidth}
  \centering
  \includegraphics[width=0.8\textwidth]{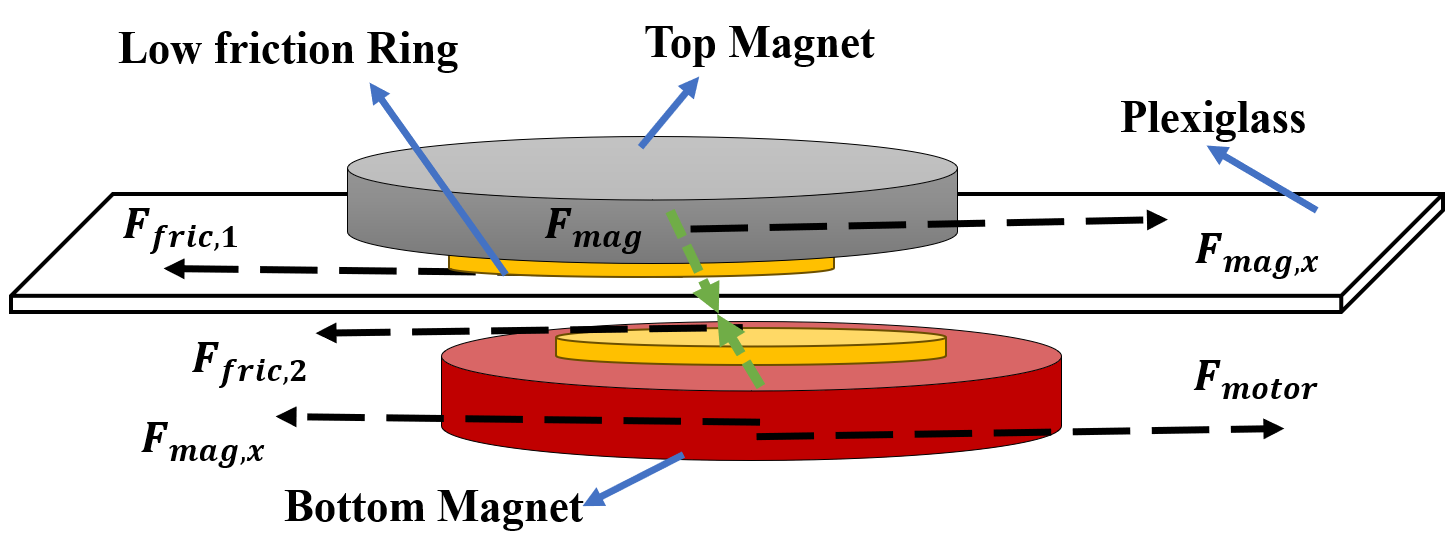} % Width set to 80% of minipage width
  \caption{Forces acted upon magnets  when going to the right}
\label{system}
\end{minipage}

% First subfigure with specific width and label above
\begin{minipage}{.48\textwidth}
  \centering
  \includegraphics[width=0.8\textwidth]{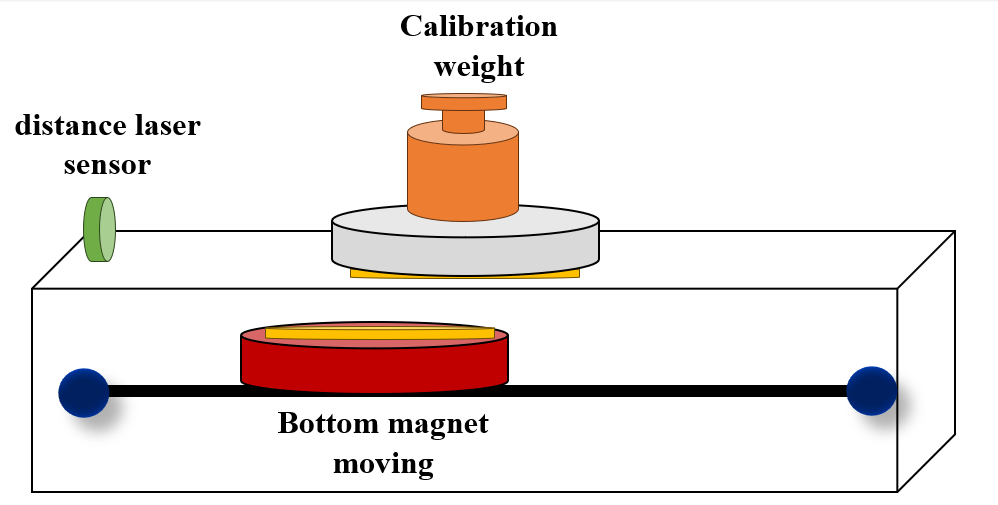} % Width set to 80% of minipage width
  \centerline{(a)}
\end{minipage}

\vspace{0.05em} % Adds some vertical space between the subfigures

% Second subfigure with label above
\begin{minipage}{.48\textwidth}
  \centering
  \includegraphics[width=0.8\textwidth]{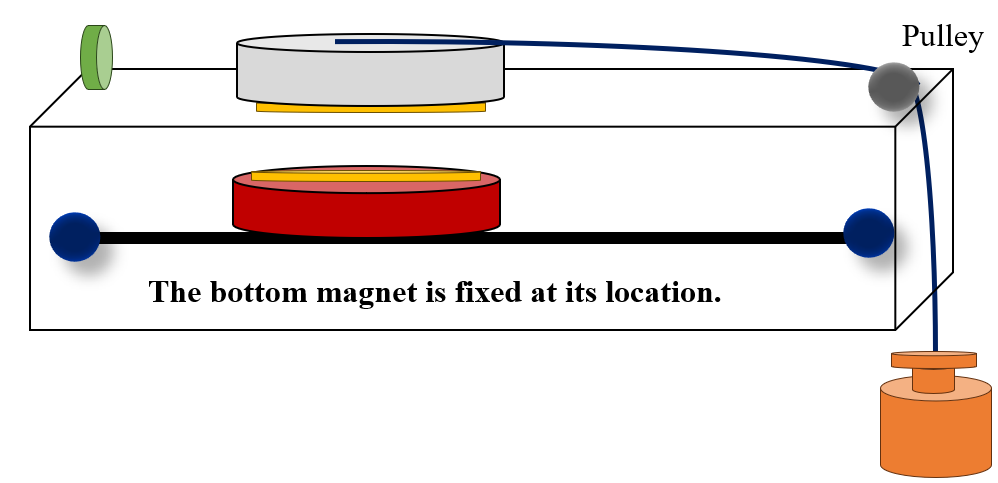} % Width set to 80% of minipage width
  \centerline{(b)}
\end{minipage}

\caption{a) The dynamic trial for measuring the offset of the magnets while the system is operating. b) The static trial for testing the strength of the magnetism field while the system is locked and not moving.}
\label{behavior}

\end{figure}

\section{Platform Development and Characterization}
\subsection{Platform Design}
\noindent Our platform operates on a single axis with a Teknic DC motor driving a belt looped around two fixed pulleys, moving a linear slider with a permanent magnet end-effector (Figure \ref{hardware}-c). Five photo interrupter sensors track the end-effector's position via an Arduino board, with end sensors acting as safety stops. Enclosed in plexiglass for safety and weight counterbalance, the mechanism ensures smooth motion with low friction rings. Technician-defined movements follow initial motor positioning via a GUI control panel.  Our system's GUI, shown in Figure \ref{hardware} (panels a and b), is developed in Python 3.11 for visual feedback and technician control. Featuring a turtle icon as a cursor for the end-effector's location, it allows setting and adjusting the end-effector's path. PySerial manages hardware communication. The GUI supports inputting user and trial data, defining trajectories, and visualizing trial data for technician analysis.

\subsection{Nonlinear dynamic Model of the System}
\noindent In this case study, we focused on cylindrical permanent magnets. While usually analyzed as magnetic dipoles, our specific magnet shapes and distances required a detailed analysis. Figure \ref{system} shows a free-body diagram of our novel actuating mechanism and the forces on the magnets.
The following formula calculates the attractive magnetostatic force between two laterally displaced cylindrical magnets according to \cite{vokoun2009magnetostatic}:

\begin{equation}
A = \frac{1}{d^2} + \frac{1}{(d+2h)^2} - \frac{2}{(d+h)^2}
\end{equation}

\noindent \begin{equation}
    F_{\text{mag}, x} = -\frac{\pi K_d R^4}{2} \Bigg[ A \\
 - \frac{3}{2}(p_1 - p_2)^2 A^2 \Bigg] \cdot \text{atan}(\alpha) 
\end{equation}

\noindent \( p_1 \) and \( p_2 \) are the positions of the bottom and top magnets. \(alpha\) is the angle between the magnetic force and the x-axis.
\noindent As for the friction analysis, for the top magnet, which exhibits a stick-slip motion, the friction is best described using the Stribeck model according to \cite{liu2015experimental}:

\begin{equation}
   F_{\text{fric},1} = \left[F_c + (F_s - F_c) \cdot e^{-\left(\frac{v}{v_s}\right)^2}\right] \cdot \text{sgn}(v) + K_v \cdot v
\end{equation}
Table \ref{magparam} shows values and units of these parameters.

\noindent\begin{table}[!ht]
    \centering
    \caption{Formula (1), (2), and (3) Parameters Details}
    \begin{tabular}{|c|c|c|c|} \hline 
         Parameter&  Value &  Unit & Description\\ \hline 
         $d$ &  0.007 &   m &  Separation distance\\ \hline 
 $h$ & 0.0127 & m & Magnet thickness\\ \hline
 $R$ & 0.0762 & m & Radius\\\hline
 $\mu_{c2}$ & 0.01 & N.s/m & Kinetic damping coefficient\\\hline
 $\mu_{s2}$ & 0.1 & N.s/m & Static damping coefficient\\\hline
 $v_s$ & 0.08 & m/s & Stribeck velocity\\\hline
 $K_{v}$ & 0.01 & N.s/m & Viscous damping coefficient \\\hline
 $K_d$ & 49.3 & $N/m^2$ & Magnetostatic energy constant\\\hline
    \end{tabular}
    \label{magparam}
\end{table}

\noindent The bottom magnet, which changes direction without stick-slip motion, primarily experiences viscous friction. Due to the motor force, the minimal static friction with the glass surface becomes negligible. The system is modeled as a two-degree-of-freedom (2-DOF) system, with magnet kinematics as outputs and motor force as input. The top magnet's displacement correlates with the bottom magnet's displacement plus an offset.
The net forces for the bottom magnet (\( F_{\text{net},1} \)) and the top magnet (\( F_{\text{net},2} \)) are calculated as:

\begin{equation}
    F_{\text{net},1} = F_{\text{motor}} - F_{\text{mag}, x} -     F_{\text{fric},1}
\end{equation}

\begin{equation}
  F_{\text{net},2} = F_{\text{mag}, x} - F_{\text{fric},2}
\end{equation}

Here, \( F_{\text{motor}} \) is the motor force. Parameters were manually selected and adjusted using benchmarks from similar studies.

\begin{figure*}[b]
\centering
\includegraphics[width=0.78\textwidth]{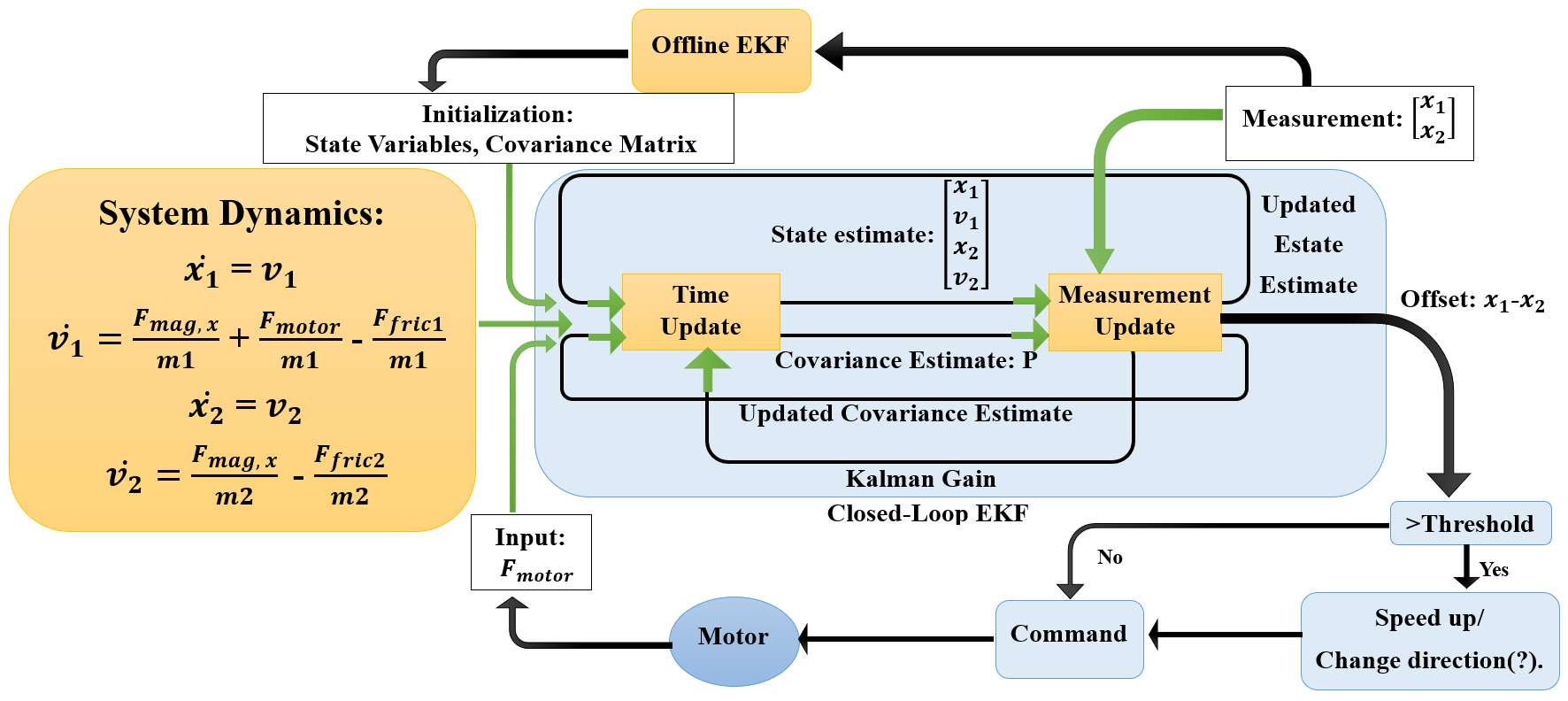}
\caption{Block diagram of the algorithm development with Extended Kalman Filter for the observable system. $x_{1}$, $x_{2}$, $v_{1}$, and $v_{2}$ are the positions and the velocities of the bottom and the top magnet respectively}
\label{diagram}
\end{figure*}

\subsection{Static and Dynamic Characterization}

\noindent Our investigation focused on the synchronized movement of cylindrical permanent magnets within a rehabilitation platform (Figure \ref{behavior}). The aim was to ensure the magnets moved in unison and remained attached under various patient forces. 

In dynamic testing (Figure \ref{behavior}-a), we used a GHB38 rotary encoder to track the bottom magnet's motion and a HiLetgo VL53L0X sensor to monitor the top magnet's movement. This setup simulated patient interaction and ensured movement synchronization, critical for therapeutic effectiveness.

In static testing (Figure \ref{behavior}-b), we fixed the bottom magnet and applied different forces to the top magnet using calibration weights. This established the threshold force for potential detachment, ensuring it was high enough to prevent accidental detachment but low enough to avoid excessive resistance.

We considered only transitional and normal forces, as encountered in passive rehabilitation, but also recognized potential additional resistance from patients with limited limb functionality. This ensured our system could maintain efficacy and safety in real-world use.

\section{Algorithm Development}
\noindent To investigate the system and prevent magnet detachment due to unsolicited forces, we developed a resilient closed-loop algorithm with Extended Kalman Filter (EKF) (Figure \ref{diagram}). This algorithm monitors and manages the distance between the magnets, ensuring smooth and uninterrupted patient interaction. 

\subsection{state-space representation}

\noindent To simplify the system dynamics (Figure \ref{diagram}, left side) we use state-space representation, a mathematical model that describes a system with input, output, and state variables related by differential equations\cite{terejanu2008extended}. This involves two equations:

1. Describe how the system's state (positions and velocities of the magnets, \(x\)) evolves based on the current state and input (motor force).

2. Output Equation: Relates the current state to the system's output.

For our system, the four state variables are the positions and velocities of the magnets. Our output is the position of one or both magnets. The state equations are:
\noindent
\begin{equation}
    \Dot{x}_{\text{1}} = v_{\texttt{1}} \quad \quad \Dot{x}_{\text{2}} = v_{\texttt{2}}  
\end{equation}

\noindent
\begin{align}
    \Dot{v}_{\text{1}} &= \frac{F_{\text{motor}}}{m_{\text{1}}} - \frac{F_{\text{mag,x}}}{m_{\text{1}}} - \frac{F_\text{fric,1}}{m_\text{1}} \\
    \Dot{v}_{\text{2}} &= \frac{F_{\text{mag,x}}}{m_{\text{2}}} - \frac{F_\text{fric,2}}{m_\text{2}}
\end{align}

\subsection{Closed-loop state estimation using EKF}

\noindent In our setup, the Extended Kalman Filter (EKF, for more information, refer to \cite{ribeiro2004kalman}) was employed to ascertain the positions and velocities of the two magnets. We used the EKF algorithm with positions measured by a laser distance sensor (top magnet) and an encoder (bottom magnet) as measurements, the motor force as the control input, and the positions and velocities as state variables.  This way the system was fully observable and allowed the EKF to correct inaccuracies in the sensor data and validate the dynamic model. Due to the EKF's effectiveness, we considered phasing out the laser sensor for a sensorless estimation. Consequently, we used only the position of the bottom magnet as our measurement. Finally, we used Root Mean Square Error (RMSE) for assessing the accuracy of EKF estimations\cite{konatowski2016comparison}.

\section{Experimental Protocol}

\begin{figure*}[!ht]
   \centering
    \includegraphics[width=0.94\textwidth]{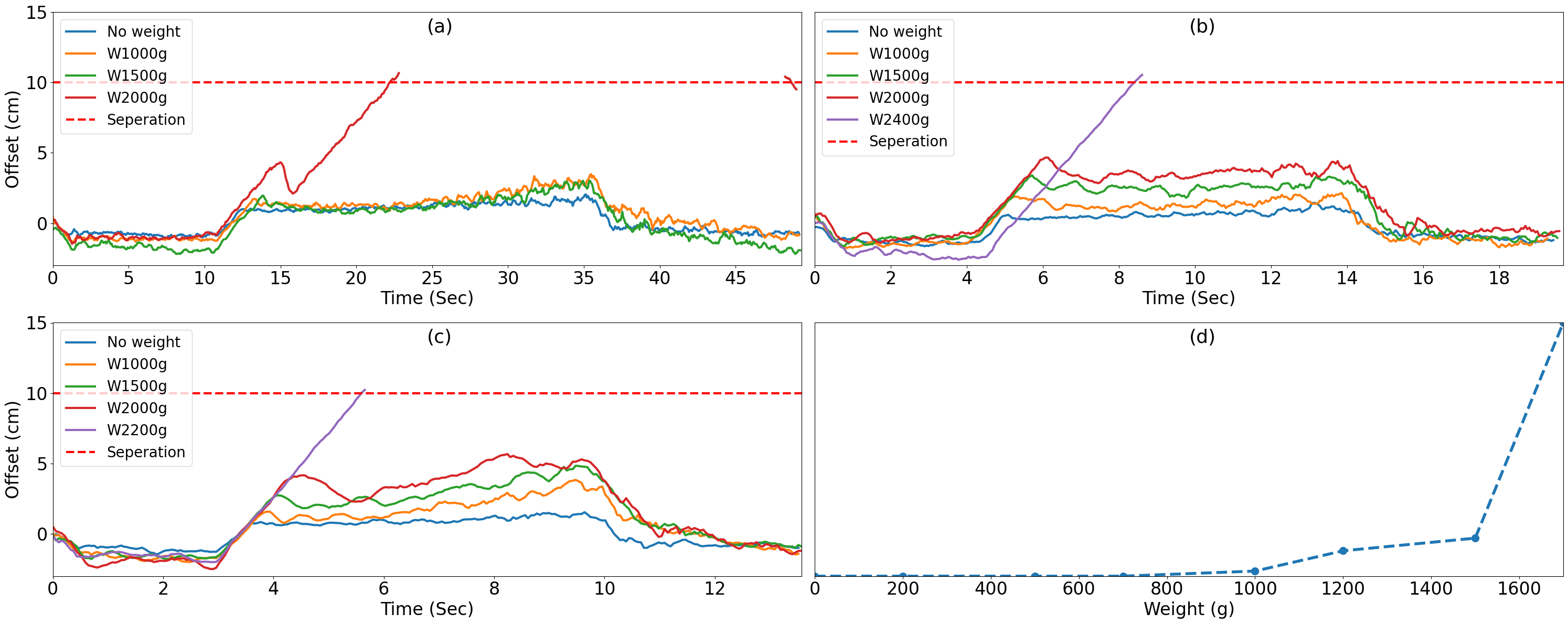}
    
    \caption{Characterization of the dynamic behavior of the system at (a) 10, (b) 20, and (c) 30 RPM speeds (1.46, 3.5, and 4.3 cm/s respectively); and (d) the static trial. During the dynamic trial, the system traveled 60 cm at three different speeds with varying calibration weights placed on the armrest on each run.}
    \label{trials}
% \end{figure*}

% \begin{figure*}[ht]
   \centering
    \includegraphics[width=0.96\textwidth]{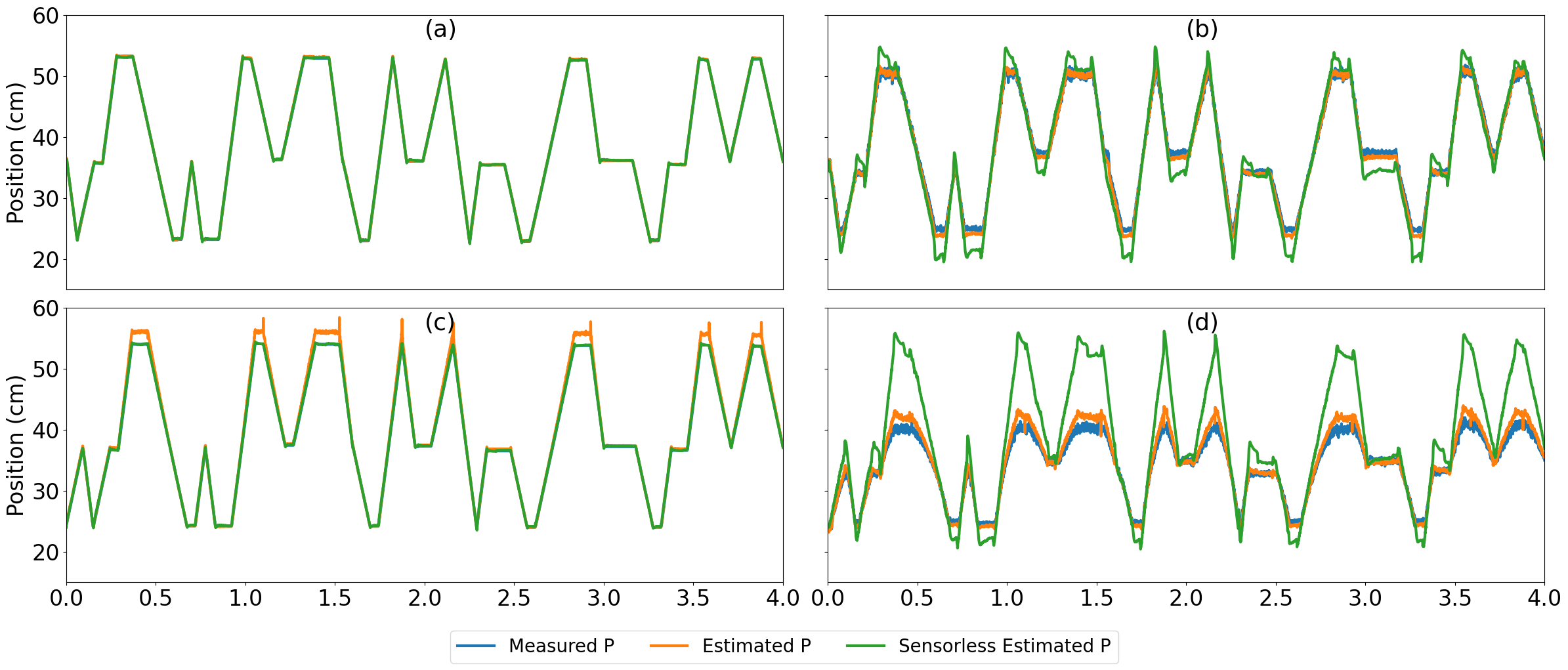}
    
    \caption{Panel a) and b) shows the measured and estimated positions using EKF and sensorless EKF of the bottom magnet (a) and the top magnet (b), with the human hand involved at two 15 (2.2 cm/s ) and 25 (3.7 cm/s) RPM speeds for one participant. Panel c) and d) display the same trajectory for one human subject when we block the laser sensor readings.}
    \label{estimation}
\end{figure*}

\noindent As a feasibility test for our system, we conducted a calibration protocol to confirm platform functionality with human interaction before data collection. We assessed participants' physical capabilities and hand range of motion, making necessary adjustments to chair and desk heights and spacing. Participants were seated comfortably with an upright posture. Once adjusted, the device moved the participant's hand along a set path with speeds of 15 and 25 RPMs (2.2 and 3.7 cm/s). We ran the exercise ran 2 times, and in between sessions, the participants filled out a survey. The entire process was under one hour.

\subsection{Participants}
\noindent We enlisted 12 healthy volunteers from the University of Rhode Island, including 9 males and 3 females aged 22 to 36 years (10 right-handed, 2 left-handed). None had a history of neurological disorders. Before starting, each participant received a detailed explanation of the experiment and gave written consent. The Institutional Review Board (IRB) approved the research methods to ensure they met ethical research standards.

\section{Results}

\begin{figure}[!ht]
\centering
\includegraphics[width=0.46\textwidth]{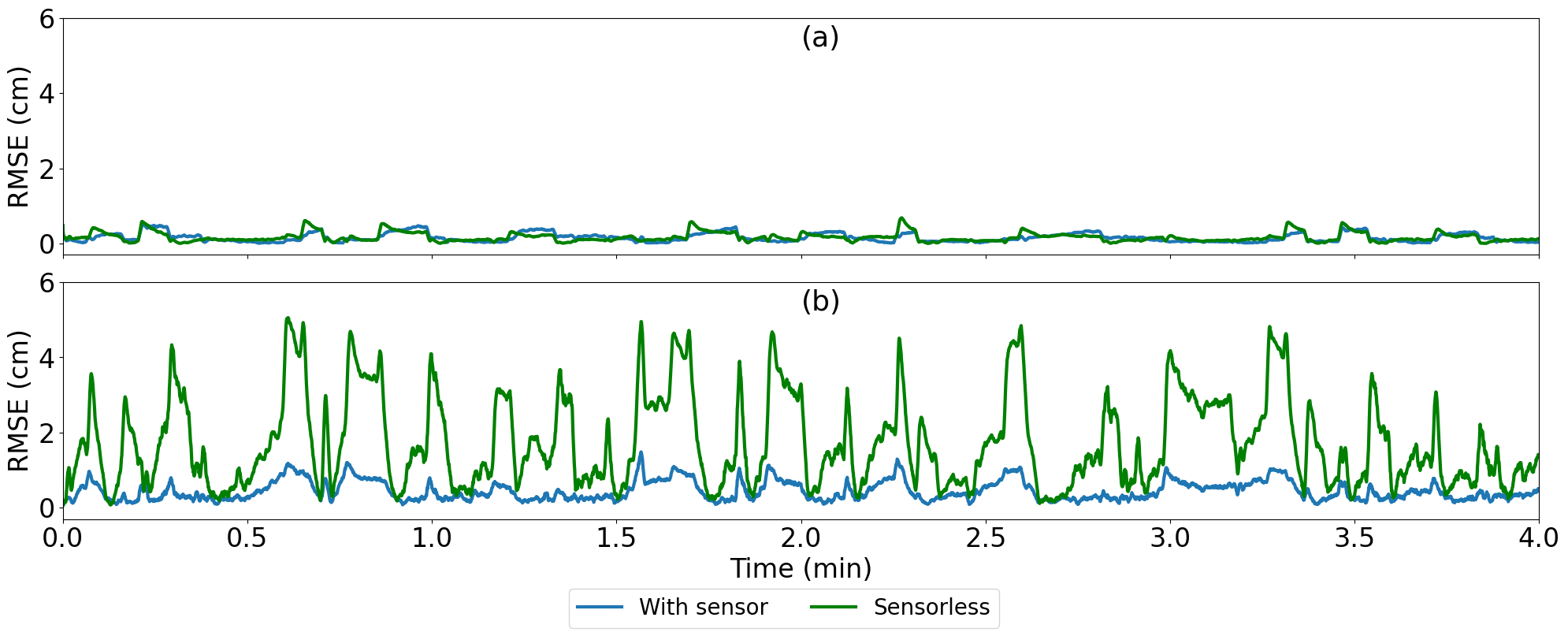}
\caption{RMSE plots for the bottom (a) and the top magnet (b)  from the two EKF estimations for one participant. }
\label{rmse}
\centering
\includegraphics[width=0.46\textwidth]{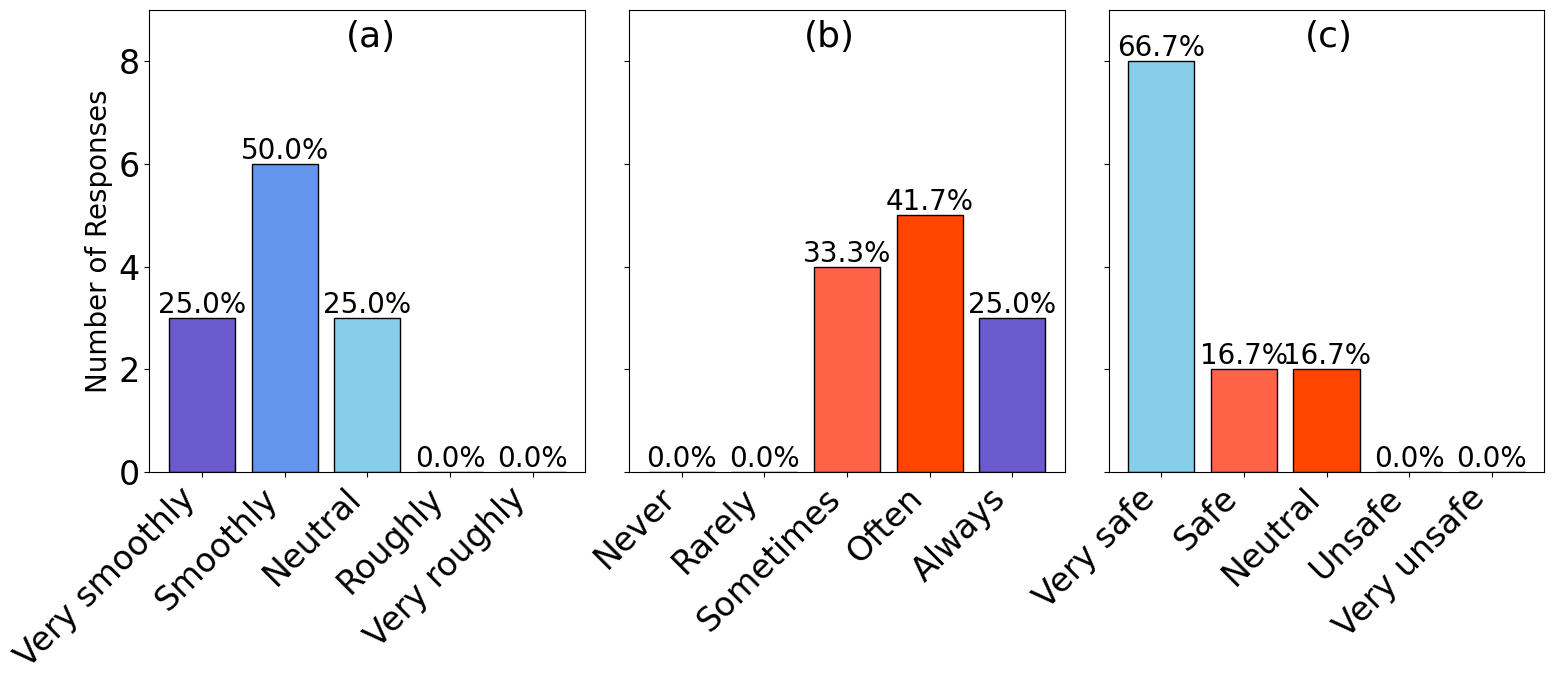}
\caption{User feedback on the performance of a magnetic-based robotic rehabilitation system in a) smoothness of magnetic interface, b) weight compensation by the system, and c) sense of safety (plexiglass).}
\label{survey}
\end{figure}

\begin{figure}[b]
\centering
\includegraphics[width=0.46\textwidth]{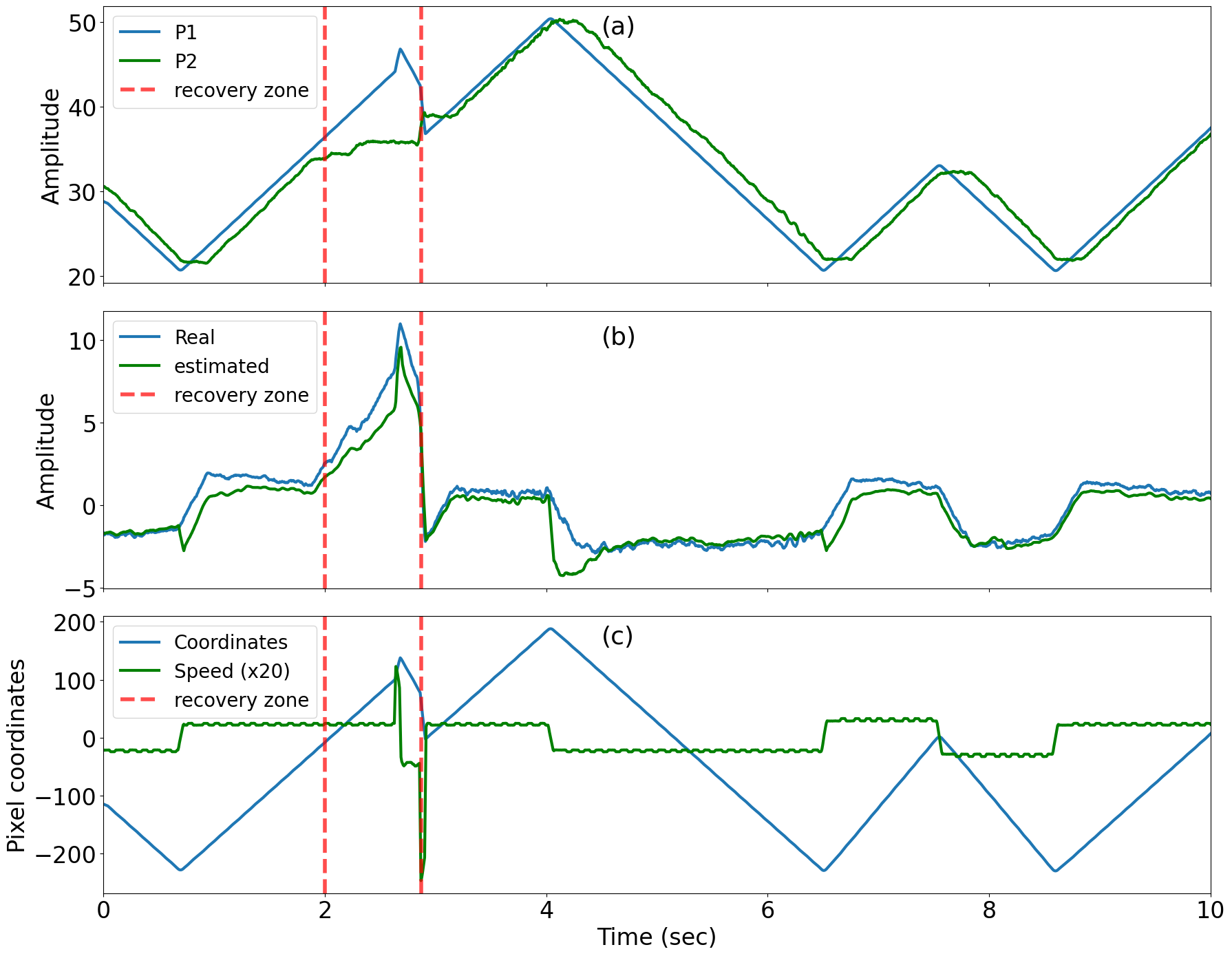}
\caption{Offset recovery between magnet positions a) bottom and top magnet positions (cm), b) Real and estimated offset (cm), and c) Turtle coordinates and speed.}
\label{recovery}
\end{figure}

\noindent Figure \ref{trials} shows the offset behavior between two magnets in a drive-follow setup, examining the impact of different weights on the detachment threshold at various speeds. At 10 RPM, heavier weights increase the amplitude, suggesting higher detachment risk. At 20 RPM, the effect of weight on detachment is less pronounced. At 30 RPM, the heaviest weight causes a significant peak, indicating a critical detachment point. According to \cite{FERRE2023R758}, a 2-kg weight pulling force is sufficient. The static test showed that the magnets held up to 1000 grams, with a complete detachment of 1700 grams.

Figure \ref{estimation} illustrates the path taken by the participants over approximately 4 minutes, with speeds of 15 RPM (2.2 cm/s) and 25 RPM (3.7 cm/s). The maximum distance covered was 30 centimeters, and they traversed approximately 4 meters. The figure also shows the distances recorded by the encoder (associated with the bottom magnet/driver: P1) and the laser distance sensor (linked to the top magnet/follower: P2). Figures \ref{estimation}-a and \ref{estimation}-b depict the EKF predictions for the magnet positions in two scenarios: with and without a sensor for 12 human subjects (both magnets follow each other through the same trajectory with a small offset in their positions). Figures \ref{estimation}-c and \ref{estimation}-d demonstrate a situation where we intentionally blocked the laser sensor reading to test our sensorless EKF estimations. We observe that when the system loses visual tracking of the top magnet, the real and estimated measurements from the first EKF become significantly inaccurate. However, the sensorless EKF successfully continues tracking the position of the top magnet. This scenario highlights the importance of sensorless estimation in preventing system failure when sensor readings are compromised. The model effectively predicts the top magnet's position despite some spikes when the bottom magnet interacts with the photo interrupter sensors. These spikes are likely due to mechanical disturbances at the start of the movement, causing the EKF to perform less optimally, particularly in the sensorless system.

Figure \ref{rmse} shows the root mean square error (RMSE) between the real and estimated positions for the bottom (top panel) and top magnet (bottom panel) during a 4-minute trial, comparing sensor-based and sensorless Extended Kalman Filter (EKF) estimations. The sensor-based method consistently shows lower errors, especially for the top magnet, where sensorless estimations exhibit higher fluctuations. Table \ref{rmsetab} provides the average RMSE values for 12 participants. A paired t-test showed no significant difference for the bottom magnet (p = 0.339), indicating minimal sensor impact. However, for the top magnet, the sensor significantly improved RMSE ($p < 0.001$), with the sensor-based EKF achieving a much lower RMSE (0.44 cm) compared to sensorless EKF (1.7 cm).

\begin{table}
    \centering
    \caption{average of RMSE values of the estimated top and bottom magnet positions for 12 participants during the trial.}
    \begin{tabular}{|c|c|c|c|} \hline 
         Magnet&  RMS (sensor-based)&  RMS (sensorless)& P\_value\\ \hline 
         Bottom&  0.157&  0.168& $>0.05$\\ \hline 
         Top&  0.44&  1.7& $< 0.001$\\ \hline
    \end{tabular}
    \label{rmsetab}
\end{table}

Figure \ref{survey} presents user feedback on three aspects of a magnetic-based robotic rehabilitation system. Most users rated the interface as operating "Smoothly" or "Very Smoothly" (75\%), while 66.7\% felt "Very Safe" with the plexiglass setup. Additionally, 66.7\% noted effective weight compensation "Often" or "Always." Kruskal-Wallis tests revealed significant differences in responses ($p < 0.01$).  

Figure \ref{recovery} demonstrates the performance of the proportional controller in recovering magnet detachment during perturbations. Figure \ref{recovery}-a shows the positions of the bottom and top magnets (P1 and P2), with a brief detachment followed by recovery within the marked zone. Figure \ref{recovery}-b highlights the real and estimated offset, where the controller quickly minimizes the error after detachment. Figure \ref{recovery}-c shows the turtle's coordinates and speed, with a brief spike during detachment that stabilizes in the recovery zone. This illustrates the controller's effectiveness in maintaining stability after perturbations.

\section{Discussion, Limitations, and Future Work}
\noindent We developed a novel actuation mechanism utilizing magnets for upper limb rehabilitation, focusing on enhancing safety and smoothness of movement.  While magnetic technologies have been explored in other rehabilitation studies, our approach uniquely targets end-effector rehabilitation robots, which has not been addressed in similar research \cite{Chaloupka2002, brainsci12010113, abbott2020magnetic}. As a proof of concept, we validated our integrated framework through a combination of complex experimental and theoretical work in 1D translational motion. Our system demonstrated that magnets could move in unison with smooth, synchronized motion. We verified this behavior by analyzing both static and dynamic system responses (without a human in the loop), confirming that it has the necessary magnet power, durability, and compliance for secure human interaction. To mitigate potential disturbances from the environment and the patient, we implemented a closed-loop Extended Kalman Filter (EKF), enhancing the system's adaptability. The EKF continuously monitored magnet alignment and ensured accurate position and velocity estimates, even in the absence of direct sensing of the upper magnet. By leveraging the lower magnet's position, our design provided reliable and responsive tracking, enhancing precision without the need for complex hardware like high-resolution encoders. Our platform design effectively compensates for the weight of the hand, wrist, and arm, reducing joint strain and minimizing the risk of damage. This is a key improvement over other systems, where users are required to bear the full weight of their hand on the platform, which can lead to excessive strain \cite{campolo2014h, wu2019adaptive}. Our system focuses on shoulder rehabilitation while excluding wrist involvement. The minor rotational movement of the magnets restricted wrist motion and maintained a stable joint angle, which we verified through detailed analysis. Notably, participant feedback from surveys indicated that the system is safe, smooth, and effective. We implemented a basic proportional controller as a proof of concept to manage disruptions from spasticity in post-stroke patients during operations. This controller restores the alignment between magnets if they become detached due to sudden perturbations. However, integrating more advanced control algorithms could significantly enhance the system’s responsiveness, particularly in assistive rehabilitation scenarios where patient and robot interact more dynamically \cite{basteris2014training}. In such applications, the inclusion of electromagnets and advanced control algorithms would allow for finer and more responsive motion control, crucial for tasks that require precise, subtle adjustments. Overall, while the system was originally designed to assist patients with severe motor impairments, its design allows for customization across different stages of rehabilitation, further expanding its potential applications.

\section{conclusion}
Our research introduced a novel magnetic-based actuation mechanism for end-effector-based robotic rehabilitation platforms. By leveraging a dynamic model that incorporates frictional and magnetostatic forces, we achieved smooth and non-contact force transmission, ensuring safer and more intuitive patient interaction. The integration of the Extended Kalman Filter (EKF) for precise position tracking further enhanced the system's reliability, even allowing for sensorless operation during disturbances. Human trials demonstrated the system's effectiveness in delivering smooth operation, increased comfort, and weight compensation, with participants reporting high satisfaction levels.  These findings highlight the potential of magnetic actuation technology to revolutionize robotic rehabilitation, offering a promising avenue for further advancements in patient care and recovery. 

\section{Acknowledgment}
The research reported in this presentation was supported by the URI Foundation Grant on Medical Research. Additionally, the project was supported by the Rhode Island INBRE program from the National Institute of General Medical Sciences of the National Institutes of Health under grant number P20GM103430, and by the National Science Foundation under award ID 2245558. We would also like to extend our thanks to Dr. Yalda Shahriari and Dr. Mariusz Furmanek for their invaluable guidance, support, and contributions throughout the course of this project.

\bibliographystyle{IEEEtran}
\bibliography{main}

\end{document}